\documentclass[10pt]{article}

\usepackage[utf8]{inputenc}

\usepackage{hyperref}
\usepackage{verbatim}

\usepackage{graphicx}
\usepackage{caption}
\usepackage{subcaption}
\usepackage{amssymb}
\usepackage{ textcomp }
\usepackage{color}
\usepackage{natbib}
\usepackage{enumerate}
\usepackage{epstopdf}
\usepackage{amsopn}
\usepackage{amsmath}
\usepackage{amsthm}
\usepackage{multicol}

\usepackage{url}

\usepackage{amsmath,amsthm,amssymb}

\newtheorem{thm}{Theorem}[section]

\theoremstyle{definition}
\newtheorem{defi}[thm]{Definition}

\numberwithin{equation}{section}

\setcitestyle{numbers,open={[},close={]}}

\newcommand{\bitem}{\begin{itemize}}
\newcommand{\eitem}{\end{itemize}}
\newcommand{\beq}{\begin{equation}}
\newcommand{\eeq}{\end{equation}}

\newcommand{\ip}[2]{\langle#1,#2\rangle}
\newcommand{\norm}[1]{\|#1\|}
\DeclareMathOperator*{\argmin}{argmin}

\newcommand{\R}{\mathbb{R}}
\newcommand{\N}{\mathbb{N}}
\newcommand{\Z}{\mathbb{Z}}

\def\cE{{\mathcal{E}}}
\def\cR{{\mathcal{R}}}

\newcommand{\cNN}{\mathcal{NN}}

\title{The Mathematics of Artificial Intelligence}

\author{Gitta Kutyniok}

\date{}

\begin{document}

\maketitle

\begin{abstract}
We currently witness the spectacular success of artificial intelligence
in both science and public life. However, the development of a rigorous
mathematical foundation is still at an early stage. In this survey article,
which is based on an invited lecture at the International Congress of Mathematicians 2022,
we will in particular focus on the current ``workhorse'' of artificial
intelligence, namely deep neural networks. We will present the main theoretical
directions along with several exemplary results and discuss key
open problems.
\end{abstract}

\section{Introduction}

Artificial intelligence is currently leading to one breakthrough after the other, both in
public with, for instance, autonomous driving and speech recognition, and in the sciences
in areas such as medical diagnostics or molecular dynamics. In addition, research on artificial intelligence
and, in particular, on its theoretical foundations, is progressing at an unprecedented rate.
One can envision that according methodologies will in the future drastically change the
way we live in numerous respects.

\subsection{The Rise of Artificial Intelligence}

Artificial intelligence is however not a new phenomenon.
In fact, already in 1943, McCulloch and Pitts started to develop algorithmic approaches to
learning by mimicking the functionality of the human brain, through artificial neurons which
are connected and arranged in several layers to form artificial neural networks. Already at
 that time, they had a vision for the implementation of artificial intelligence. However, the community did not
fully recognize the potential of neural networks. Therefore, this first wave of artificial
intelligence was not successful and vanished.  Around 1980, machine learning became popular
again, and several highlights can be reported from that period.

The real breakthrough and with it a new wave of artificial intelligence came around 2010
with the extensive application of \emph{deep} neural networks. Today, this model might be considered the
``workhorse'' of artificial intelligence, and in this article we will focus predominantly
on this approach. The structure of deep neural networks is precisely the structure
McCulloch and Pitts introduced, namely numerous consecutive layers of artificial neurons.
Today two main obstacles from previous years have also been eliminated; due to the drastic improvement of
computing power the training of neural networks with hundreds of layers in the sense of
\emph{deep} neural networks is feasible, and we are living in the age of data, hence vast
amounts of training data are easily available.

\subsection{Impact on Mathematics}

The rise of artificial intelligence also had a significant impact on various fields of mathematics.
Maybe the first area which embraced these novel methods was the area of inverse problems, in
particular, imaging science where such approaches have been used to solve highly ill-posed problems such as denoising,
inpainting, superresolution, or (limited-angle) computed tomography, to name a few. One
might note that due to the lack of a precise mathematical model of what an image is, this
area is particularly suitable for learning methods. Thus, after a few years, a
change of paradigm could be observed, and novel solvers are typically at least to
some extent based on methods from artificial intelligence. We will discuss further details
in Subsection \ref{subsec:AIfMP_IP}.

The area of partial differential equations was much slower to embrace these new techniques,
the reason being that it was not per se evident what the advantage of methods from artificial
intelligence for this field would be. Indeed, there
seems to be no need to utilize learning-type methods, since a partial differential equation
is a rigorous mathematical model. But, lately, the observation that deep neural networks are able to beat the curse of dimensionality in high
dimensional settings led to a change of paradigm in this area as well. Research at
the intersection of numerical analysis of partial differential equations and artificial
intelligence therefore accelerated since about 2017. We will delve further into this topic in
Subsection \ref{subsec:AIfMP_PDE}.

\subsection{Problems of Artificial Intelligence}

However, as promising as all these developments seem to be, a word of caution is required.
Besides the fact that the practical limitations of methods such as deep neural networks have not been
explored at all and at present neural networks are still considered a ``jack of all trades'', it is
even more worrisome that a comprehensive theoretical foundation is completely lacking.
This was very prominently stated during the major conference in artificial intelligence and
machine learning, which is NIPS (today called NeurIPS) in 2017, when Ali Rahimi from Google
received the Test of Time Award and during his plenary talk stated that ``Machine learning
has become a form of alchemy''. This raised a heated discussion to which extent a
theoretical foundation does exist and is necessary at all. From a mathematical viewpoint,
it is crystal clear that a fundamental mathematical understanding of artificial intelligence
is inevitably necessary, and one has to admit that its development is currently in a preliminary state at best.

This lack of mathematical foundations, for instance, in the case of deep neural networks, results
in a time-consuming search for a suitable network architecture, a highly delicate trial-and-error-based (training)
process, and missing error bounds for the performance of the trained neural network. One
needs to stress that, in addition, such approaches also sometimes unexpectedly fail dramatically
when a small perturbation of the input data causes a drastic change of the output leading to
radically different\textemdash and often wrong\textemdash decisions. Such adversarial
examples are a well-known problem, which becomes severe in sensitive applications such as
when a minor alterations of traffic signs, e.g, the placement of stickers, causes autonomous
vehicles to suddenly reach an entirely wrong decision. It is evident that such
robustness problems can only be tackled by a profound mathematical approach.

\subsection{A Need for Mathematics}

These considerations show that there is a tremendous need for mathematics in the area of
artificial intelligence. And, in fact, one can currently witness that numerous mathematicians
move to this field, bringing in their own expertise. Indeed, as we will discuss in
Subsection \ref{subsec:mainthreads}, basically all areas of mathematics are required to
tackle the various difficult, but exciting challenges in the area of artificial intelligence.

One can identify two different research directions at the intersection of mathematics and
artificial intelligence:
\bitem
\item {\em Mathematical Foundations for Artificial Intelligence}. This direction aims for
deriving a deep mathematical understanding. Based on this it strives to overcome current
obstacles such as the lack of robustness or places the entire training process on solid
theoretical feet.
\item {\em Artificial Intelligence for Mathematical Problems}. This direction focuses on
mathematical problem settings such as inverse problems and partial
differential equations with the goal to employ methodologies from artificial intelligence
to develop superior solvers.
\eitem

\subsection{Outline}

Both research directions will be discussed in this survey paper, showcasing some novel
results and pointing out key future challenges for mathematics. We start with an
introduction into the mathematical setting, stating the main definitions and notations
(see Section~\ref{sec:mathematicalsetting}). Next, in Section \ref{sec:MFfAI}, we delve
into the first main direction, namely mathematical foundations for artificial intelligence,
and discuss the research threads of expressivity, optimization, generalization, and explainability.
Section \ref{sec:AIfMP} is then devoted to the second main direction, which is artificial
intelligence for mathematical problems, and we highlight some exemplary
results. Finally, Section \ref{sec:conclusion} states the seven main mathematical problems
and concludes this article.

%
%

\section{The Mathematical Setting of Artificial Intelligence}
\label{sec:mathematicalsetting}

We now get into more details on the precise definition of a deep neural network,
which is after all a purely mathematical object. We will also touch upon the typical application
setting and training process, as well as on the current key mathematical directions.

\subsection{Definition of Deep Neural Networks}

The core building blocks are, as said, artificial neurons. For their definition, let us recall
the structure and functionality of a neuron in the human brain. The basic elements of such a neuron
are dendrites, through which signals are transmitted to its soma while being scaled/amplified due
to the structural properties of the respective dendrites. In the soma of the neuron, those incoming
signals are accumulated, and a decision is reached whether to fire to other neurons or not, and
also with which strength.

This forms the basis for a mathematical definition of an artificial neuron.

\begin{defi}
An \emph{artificial neuron} with \emph{weights} $w_1,..., w_n \in \mathbb{R}$,
\emph{bias} $b \in \mathbb{R}$, and \emph{activation function}
$\rho: \mathbb{R} \rightarrow \mathbb{R}$ is defined as the function $f: \mathbb{R}^n \rightarrow \mathbb{R}$ given by
\[
f(x_1,...,x_n) = \rho\left(\sum_{i=1}^n x_i w_i - b\right) = \rho(\langle x, w \rangle - b),
\]
where $w=(w_1,...,w_n)$ and $x=(x_1,...,x_n).$
\end{defi}

By now, there exists a zoo of activation functions with the most well-known ones being as follows:
\begin{itemize}
\item[(1)] Heaviside function $\rho (x) = \begin{cases} 1, & x>0, \\
0, & x \leq 0.
\end{cases}$
\item[(2)] Sigmoid function $\rho(x) = \frac{1}{1+e^{-x}}$.
\item[(3)] Rectifiable Linear Unit (ReLU) $\rho(x) = \max \{ 0, x \}$.
\end{itemize}
We remark that of these examples, the by far most extensively used activation function
is the ReLU due to its simple piecewise linear structure, which is advantageous in the training
process and still allows superior performance.

Similar to the structure of a human brain, these artificial neurons are now being concatenated
and arranged in layers, leading to an (artificial feed-forward) neural network. Due to the
particular structure of artificial neurons, such a neural network consists of compositions of
affine linear maps and activation functions.
Traditionally, a deep neural network is then defined as the resulting function. From a mathematical
standpoint, this bears the difficulty that different arrangements lead to the same function.
Therefore, sometimes a distinction is made between the architecture of a neural network and the
corresponding realization function (see, e.g., \cite{MMDL}). For this article, we will however
avoid such technical delicacies and present the most standard definition.

\begin{defi} \label{defi:DNN}
Let $d \in \N$ be the dimension of the input layer, $L$ the number of layers, $N_0 := d$, $N_\ell$, $\ell = 1, \ldots, L$,
the dimensions of the hidden and last layer, $\rho : \R \to \R$ a (non-linear) activation function, and,
for $\ell = 1, \ldots, L$, let $T_{\ell}$ be the affine-linear functions
\[
T_{\ell}: \R^{N_{\ell-1}} \to \R^{N_\ell}, \qquad T_\ell x = W^{(\ell)} x + b^{(\ell)},
\]
with $W^{(\ell)} \in \mathbb{R}^{N_\ell \times N_{\ell-1}}$ being the weight matrices and $b^{(\ell)} \in \R^{N_\ell}$
the bias vectors of the $\ell$th layer. Then $\Phi: \R^d \to \R^{N_L}$, given by
\[
\Phi(x) = T_L\rho( T_{L-1}\rho( \dots \rho (T_{1}(x)))), \qquad x\in \R^d,
\]
is called \emph{(deep) neural network} of \emph{depth} $L$.
\end{defi}

\begin{figure}[h]
\centering
\includegraphics[height=4cm]{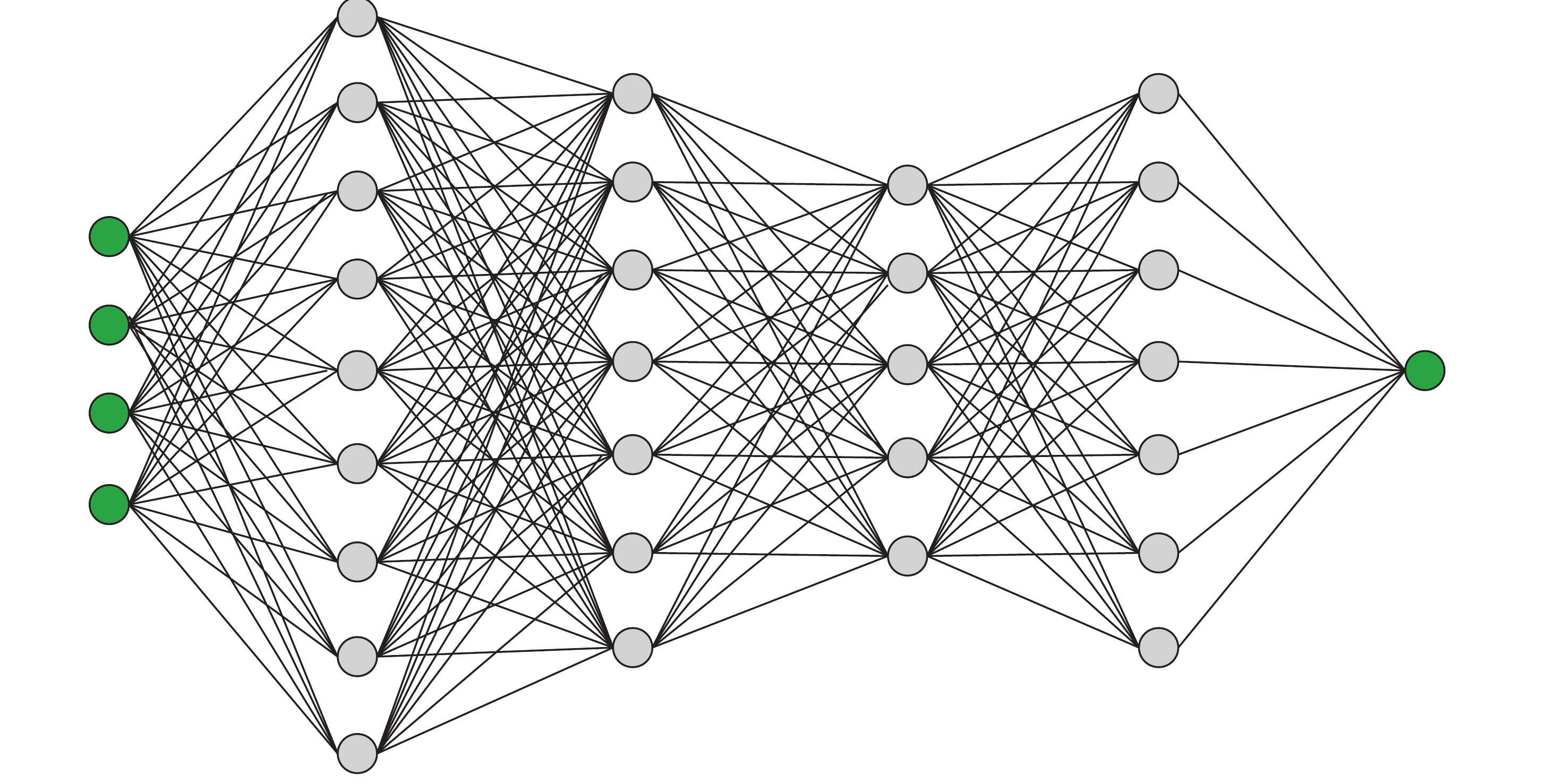}
\caption{Deep neural network $\Phi: \R^4 \to \R$ with depth $5$.}
\label{fig:2}
\end{figure}

Let us already mention at this point that the weights and biases are the free parameters which
will be learned during the training process. An illustration of the multilayered structure of a deep neural network can be found in Figure \ref{fig:2}.

\subsection{Application of a Deep Neural Network}

Aiming to identify the main mathematical research threads, we first have to understand
how a deep neural network is used for a given application setting.

\emph{Step 1 (Train-test split of the dataset):} We assume that we are given samples $(x^{(i)},y^{(i)})_{i=1}^{\widetilde{m}}$ of inputs and outputs. The
task of the deep neural network is then to identify the relation between those. For instance, in
a classification problem, each output $y^{(i)}$ is considered to be the label of the respective class to
which the input $x^{(i)}$ belongs. One can also take the viewpoint that $(x^{(i)},y^{(i)})_{i=1}^{\widetilde{m}}$ arises
as samples from a function such as $g : \mathcal{M} \to \{1,2,\ldots,K\}$, where $\mathcal{M}$
might be a lower-dimensional manifold of $\R^d$, in the sense of $y^{(i)} = g(x^{(i)})$ for all $i = 1, \ldots, \widetilde{m}$. \\
The set $(x^{(i)},y^{(i)})_{i=1}^{\widetilde{m}}$ is then split into a training data set $(x^{(i)},y^{(i)})_{i=1}^m$
and a test data set $(x^{(i)},y^{(i)})_{i=m+1}^{\widetilde{m}}$. The training data set is\textemdash  as the
name indicates\textemdash used for training, whereas the test data set will later on solely be exploited for testing the
performance of the trained network. We emphasize that the neural network is not exposed to the test
data set during the entire training process.

\emph{Step 2 (Choice of architecture):}  For preparation of the learning algorithm, the architecture of the neural network
needs to be decided upon, which means the number of layers $L$, the number of neurons in each layer
$(N_\ell)_{\ell=1}^L$, and the activation function $\rho$ have to be selected. It is known that
a fully connected neural network is often difficult to train, hence, in addition, one typically
preselects certain entries of the weight matrices $(W^{(\ell)})_{\ell=1}^L$ to already be set to
zero at this point.\\
For later purposes, we define the selected class of deep neural networks by $\cNN_{\theta}$ with
$\theta$ encoding this chosen architecture.

\emph{Step 3 (Training):}  The next step is the actual training process, which
consists of learning the affine-linear functions $(T_\ell)_{\ell=1}^L = (W^{(\ell)} \cdot + b^{(\ell)})_{\ell=1}^L$.
This is accomplished by minimizing the \emph{empirical risk}
\begin{equation}\label{eq:risk1}
\widehat{\cR}(\Phi_{(W^{(\ell)},b^{(\ell)})_\ell}) := \frac{1}{m}\sum_{i=1}^{m} (\Phi_{(W^{(\ell)},b^{(\ell)})_\ell}(x^{(i)}) - y^{(i)})^2.
\end{equation}
A more general form of the optimization problem is
\begin{equation}\label{eq:risk2}
\min_{(W^{(\ell)},b^{(\ell)})_\ell} \sum_{i=1}^{m} \mathcal{L}(\Phi_{(W^{(\ell)},b^{(\ell)})_\ell}(x_i),y^{(i)}) + \lambda \mathcal{P}((W^{(\ell)},b^{(\ell)})_\ell),
\end{equation}
where $\mathcal{L}$ is a loss function to determine a measure of closeness between the network
evaluated in the training samples and the (known) values $y^{(i)}$ and where $\mathcal{P}$ is a penalty/regularization term
to impose additional constraints on the weight matrices and bias vectors.\\
One common algorithmic approach is gradient descent. Since, however, $m$ is typically very large, this is computationally
not feasible. This problem is circumvented by randomly selecting only a few gradients in each iteration, assuming
that they constitute a reasonable average, which is coined \emph{stochastic gradient descent}. \\
Solving the optimization problem then yields a network
$\Phi_{(W^{(\ell)},b^{(\ell)})_\ell}: \R^d \to \R^{N_L}$, where
\[
\Phi_{(W^{(\ell)},b^{(\ell)})_\ell}(x) = T_L\rho( T_{L-1}\rho( \dots \rho (T_{1}(x)))).
\]

\emph{Step 4 (Testing):}  Finally, the performance (often also called generalization ability) of the trained neural
network is tested using the test data set $(x^{(i)},y^{(i)})_{i=m+1}^{\widetilde{m}}$ by analyzing whether
\[
\Phi_{(W^{(\ell)},b^{(\ell)})_\ell}(x^{(i)}) \approx y^{(i)}, \quad \mbox{for all } i=m+1, \ldots, \widetilde{m}.
\]

\subsection{Relation to a Statistical Learning Problem}
\label{subsec:statistics}

From the procedure above, we can already identify the selection of architecture, the optimization problem, and the
generalization ability as the key research directions for mathematical foundations of deep neural networks.
Considering the entire learning process of a deep neural network as a statistical learning problem reveals those three
research directions as indeed the natural ones for analyzing the overall error.

For this, let us assume that there exists a function $g : \R^d \to \R$ such that the training data
$(x^{(i)},y^{(i)})_{i=1}^{m}$ is of the form $(x^{(i)},g(x^{(i)}))_{i=1}^{m}$ and $x^{(i)} \in [0,1]^d$
for all $i=1, \ldots, m$. A typical continuum viewpoint to measure success of the training is to consider the
\emph{risk} of a function $f : \R^d \to \R$ given by
\begin{equation} \label{eq:risk}
\cR(f) := \int_{[0,1]^d} (f(x) - g(x))^2 \, dx,
\end{equation}
where we used the $L^2$-norm to measure the distance between $f$ and $g$. The error between the trained deep neural network $\Phi^0 (:= \Phi_{(W^{(\ell)},b^{(\ell)})_\ell}) \in \cNN_\theta$ and the optimal function $g$ can then be estimated by
\begin{equation} \label{eq:error}
\cR(\Phi^0) \leq  \underbrace{\left[\widehat{\cR}(\Phi^0) - \inf_{\Phi \in \cNN_\theta} \widehat{\cR}(\Phi)\right]}_{\text{Optimization error}} + \underbrace{2\sup_{\Phi \in \cNN_\theta} |  \cR(\Phi)  -\widehat{\cR}(\Phi)|}_{\text{Generalization error}} + \underbrace{\inf_{\Phi \in \cNN_\theta}\cR(\Phi).}_{\text{Approximation error}}
\end{equation}
These considerations lead to the main research threads described in the following subsection.

\subsection{Main Research Threads}
\label{subsec:mainthreads}

We can identify two conceptually different research threads, the first one being focused on developing
mathematical foundations of artificial intelligence and the second one aiming to use methodologies from
artificial intelligence to solve mathematical problems. It is intriguing to see how both have
already led to some extent to a paradigm shift in some mathematical research areas, most prominently
the area of numerical analysis.

\subsubsection{Mathematical Foundations for Artificial Intelligence}
\label{subsec:Thread1}

Following up on the discussion in Subsection \ref{subsec:statistics}, we can identify three research
directions which are related to the three types of errors which one needs to control in order to
estimate the overall error of the entire training process.
\bitem
\item \emph{Expressivity}. This direction aims to derive a general understanding whether and to which
extent aspects of a neural network architecture affect the best case performance of deep neural networks.
More precisely, the goal is to analyze the approximation error $\inf_{\Phi \in \cNN_\theta}\cR(\Phi)$ from
\eqref{eq:error}, which estimates the approximation accuracy when approximating $g$ by the
hypothesis class $\cNN_\theta$ of deep neural networks of a particular architecture. Typical
methods for approaching this problem are from applied harmonic analysis and approximation theory.
\item \emph{Learning/Optimization}. The main goal of this direction is the analysis of the training algorithm
such as stochastic gradient descent, in particular, asking why it usually converges to suitable local
minima even though the problem itself is highly non-convex. This requires the analysis of
the optimization error, which is $\widehat{\cR}(\Phi^0) - \inf_{\Phi \in \cNN_\theta} \widehat{\cR}(\Phi)$ (cf.
\eqref{eq:error}) and which measures the accuracy with which the learnt neural network $\Phi^0$ minimizes
the empirical risk \eqref{eq:risk1}, \eqref{eq:risk2}. Key methodologies for attacking such problems come from the
areas of algebraic/differential geometry, optimal control, and optimization.
\item \emph{Generalization}. This direction aims to derive an understanding of the out-of-sample error,
namely,\\ $\sup_{\Phi \in \cNN_\theta} |  \cR(\Phi)  -\widehat{\cR}(\Phi)|$ from \eqref{eq:error}, which
measures the distance of the empirical risk \eqref{eq:risk1}, \eqref{eq:risk2} and the actual risk
\eqref{eq:risk}. Predominantly, learning theory, probability theory, and statistics provide the
required methods for this research thread.
\eitem

A very exciting and highly relevant new research direction has recently emerged, coined explainability.
At present, it is from the standpoint of mathematical foundations still a wide open field.
\bitem
\item \emph{Explainability.} This direction considers deep neural networks, which are already trained,
but no knowledge about the training is available; a situation one encounters numerous times in practice.
The goal is then to derive a deep understanding of how a given trained deep neural network reaches
decisions in the sense of which features of the input data are crucial for a decision. The range
of required approaches is quite broad, including areas such as information theory or uncertainty
quantification.
\eitem

\subsubsection{Artificial Intelligence for Mathematical Problems}

Methods of artificial intelligence have also turned out to be extremely effective for mathematical
problem settings. In fact, the area of inverse problems, in particular, in imaging sciences, has
already undergone a profound paradigm shift. And the area of numerical analysis of partial
differential equations seems to soon follow the same path, at least in the very high dimensional
regime.

Let us briefly characterize those two research threads similar to the previous subsection on
mathematical foundations of artificial intelligence.
\bitem
\item \emph{Inverse Problems.} Research in this direction aims to improve classical model-based
approaches to solve inverse problems by exploiting methods of artificial intelligence. In order
to not neglect domain knowledge such as the physics of the problem, current approaches aim to
take the best out of both worlds in the sense of optimally combining model- and data-driven
approaches. This research direction requires a variety of techniques, foremost from areas such
as imaging science, inverse problems, and microlocal analysis, to name a few.
\item \emph{Partial Differential Equations.} Similar to the area of inverse problems, here the
goal is to improve classical solvers of partial differential equations by using ideas from
artificial intelligence. A particular focus is on high dimensional problems in the sense of
aiming to beat the curse of dimensionality. This direction obviously requires methods from areas
such as numerical mathematics and partial differential equations.
\eitem

\section{Mathematical Foundations for Artificial Intelligence}
\label{sec:MFfAI}

This section shall serve as an introduction into the main research threads aiming to develop a
mathematical foundation for artificial intelligence. We will introduce the problem settings,
showcase some exemplary results, and discuss open problems.

\subsection{Expressivity}

Expressivity is maybe the richest area at present in terms of mathematical results.
The general question can be phrased as follows: Given a function class/space $\mathcal{C}$ and a class of
deep neural networks $\cNN_\theta$, how does the approximation accuracy when approximating elements of
$\mathcal{C}$ by networks $\Phi \in \cNN_\theta$ relate to the complexity of such $\Phi$? Making this
precise thus requires the introduction of a complexity measure for deep neural networks. In the sequel,
we will choose the canonical one, which is the complexity in terms of memory requirements. Notice though
that certainly various other complexity measures exist. Further, recall that the $\| \cdot \|_0$-``norm''
counts the number of non-zero components.

\begin{defi}
Retaining the same notation for deep neural networks as in Definition~\ref{defi:DNN},
the \emph{complexity $C(\Phi)$} of a deep neural network $\Phi$ is defined by
\[
C(\Phi) := \sum_{\ell=1}^L \left(\|W^{(\ell)}\|_{0}+\|b^{(\ell)}\|_{0}\right).
\]
\end{defi}

The most well-known\textemdash and maybe even the first\textemdash result on expressivity is the universal
approximation theorem \cite{UAT1,UAT2}. It states that each continuous function on a compact domain can be
approximated up to an arbitrary accuracy by a shallow neural network.

\begin{thm}
Let $d \in \N$, $K \subset \R^d$ compact, $f: K \to \R$ continuous, $\rho:\R \to \R$ continuous and not a polynomial.
Then, for each $\epsilon >0$, there exist $N\in \N$ and $a_k, b_k \in \R, w_k \in \R^d$, $1 \le k \le N$, such that
 \[
\|f - \sum_{k = 1}^N a_k \rho(\ip{w_k}{\cdot} - b_k)\|_{\infty} \leq \epsilon.
 \]
\end{thm}

While this is certainly an interesting result, it is not satisfactory in several regards: It does not
give bounds on the complexity of the approximating neural network and also does not explain why
depth is so important. A particularly intriguing example for a result, which considers complexity
and also targets a more sophisticated function space, was derived in \cite{Yar17}.

\begin{thm}
For all $f \in C^s([0,1]^d)$ and $\rho(x)= \max\{0,x\}$, i.e., the ReLU,
there exist neural networks $(\Phi_n)_{n \in \N}$ with the number of layers of $\Phi_n$ being approximately
of the order of $\log(n)$ such that
\[
\|f-\Phi_n\|_\infty \lesssim C(\Phi_n)^{-\frac{s}{d}} \to 0 \quad \mbox{as } n \to \infty.
\]
\end{thm}

This result provides a beautiful connection between approximation accuracy and complexity of
the approximating neural network, and also to some extent takes the depth of the network into
account. However, to derive a result on optimal approximations, we first require a lower
bound. The so-called VC-dimension (Vapnik-Chervonenkis-dimension) (see also \eqref{eq:generalization})
was for a long time the main method for achieving such lower bounds. We will recall here a
newer result from \cite{BGKP19} in terms of the optimal exponent $\gamma^*(\mathcal{C})$ from
information theory to measure the complexity of $\mathcal{C}\subset L^2(\R^d)$. Notice that
we will only state the essence of this result without all technicalities.

\begin{thm} \label{thm:bound}
Let $d\in \N$, $\rho :\R \to \R$, and let $\mathcal{C} \subset L^2(\R^d)$. Further, let
\[
\text{\textbf Learn} : (0,1) \times \mathcal{C} \to \cNN_\theta
\]
satisfy that, for each $f \in \mathcal{C}$ and $0<\epsilon<1$,
\[
    \sup_{f \in \mathcal{C}}\|f - \text{\textbf Learn}(\epsilon, f)\|_{2} \leq \epsilon.
\]
Then, for all $\gamma < \gamma^*(\mathcal{C})$,
\[
\epsilon^{\gamma} \sup_{f\in \mathcal{C}}C(\text{\textbf Learn}(\epsilon, f)) \to \infty, \qquad \mbox{as } \epsilon \to 0.
\]
\end{thm}

This conceptual lower bound, which is independent of any learning algorithm, now allows to
derive results on approximations with neural networks, which have optimally small complexity
in the sense of being memory-optimal.
We will next provide an example of such a result, which at the same time answers another question
as well. The universal approximation theorem already indicates that deep neural networks seem to
have a universality property in the sense of performing at least as good as polynomial
approximation. One can now ask whether neural networks also perform as well as other existing
approximation schemes such as wavelets, or the more sophisticated system of
shearlets \cite{KL12}.

For this, let us briefly recall this system and its approximation properties. Shearlets are based on
parabolic scaling, i.e.,
\[
A_{2^j} = \left(\begin{array}{cc} 2^j & 0 \\ 0  & 2^{j/2}  \end{array}\right), \quad j \in \Z
\]
and $ \tilde{A}_{2^j} = $ diag$(2^{j/2}, {2^j})$ as well as changing the orientation via shearing defined by
\[
S_k =  \left(\begin{array}{cc}  1 & k \\ 0 & 1  \end{array}\right), \quad k \in \Z.
\]
(Cone-adapted) discrete shearlet systems can then be defined as follows, cf. \cite{KL11}. A faithful
implementation of the shearlet transform as a 2D and 3D (parallelized) fast shearlet transform
can be found at \url{www.ShearLab.org}.

\begin{defi}
The \emph{(cone-adapted) discrete shearlet system} $\mathcal{S}\mathcal{H}(\phi,\psi,\tilde{\psi})$ generated
by $\phi \in L^2(\mathbb{R}^2)$ and $\psi, \tilde{\psi} \in L^2(\mathbb{R}^2)$ is the union of
\[
\{\phi(\cdot-m) : m \in \Z^2\},
\]
\[
\{2^{3j/4} \psi(S_kA_{2^j}\, \cdot \, -m) : j \ge 0, |k| \leq \lceil2^{j/2}\rceil, m \in \Z^2
\},
\]
\[
\{2^{3j/4} \tilde{\psi}(S_k^T\tilde{A}_{2^j}\, \cdot \, -m) : j \ge 0, |k| \leq \lceil 2^{j/2} \rceil, m \in \Z^2 \}.
\]
\end{defi}

Since multivariate problems are typically governed by anisotropic features such as edges in images
or shock fronts in the solution of transport-dominated equations, the following suitable model class
of functions
was introduced in \cite{Donoho11}.

\begin{defi}
The set of \emph{cartoon-like functions} $\cE^2(\R^2)$ is defined by
\[
\cE^2(\R^2) = \{f \in L^2(\R^2) : f = f_0 + f_1 \cdot \chi_{B}\},
\]
where $\emptyset \neq B \subset [0,1]^2$ is simply connected with a $C^2$-curve with bounded curvature as its boundary, and
$f_i \in C^2(\R^2)$ with supp $f_i \subseteq [0,1]^2$ and $\|f_i\|_{C^2} \le 1$, $i=0, 1$.
\end{defi}


While wavelets are deficient in optimally approximating cartoon-like functions due to their isotropic
structure, shearlets provide an optimal (sparse) approximation rate up to a log-factor. The following
statement is taken from \cite{KL11}, where also the precise hypotheses can be found. Notice that the
justification for optimality is a benchmark result from \cite{Donoho11}.

\begin{thm}
Let $\phi, \psi, \tilde{\psi} \in L^2(\R^2)$ be compactly supported, and let $\hat{\psi}$, $\hat{\tilde{\psi}}$
satisfy certain decay conditions. Then $\mathcal{S}\mathcal{H}(\phi,\psi,\tilde{\psi})$
provides an \emph{optimally sparse approximation} of $f \in \cE^2(\R^2)$, i.e.,
\[
\norm{f - f_N}_2 \lesssim N^{-1} (\log N)^\frac32 \quad \mbox{as }  N \to
\infty.
\]
\end{thm}

One can now use Theorem \ref{thm:bound} to show that indeed deep neural networks are as good
approximators as shearlets and in fact as all affine systems. Even more,
the construction in the proof of suitable neural networks, which mimics best $N$-term approximations, also leads to memory-optimal neural networks.
The resulting
statement from \cite{BGKP19} in addition proves that the bound in Theorem \ref{thm:bound} is sharp.

\begin{thm}
Let $\rho$ be a suitably chosen activation function, and let $\epsilon>0$. Then, for all $f \in \cE^2(\R^2)$ and $N \in \N$,
there exists a neural network $\Phi$ with complexity $O(N)$ and activation function $\rho$ with
\[
\|f - \Phi\|_{2} \lesssim N^{-1+\epsilon}  \to 0 \quad \mbox{as } N \to \infty.
\]
\end{thm}

Summarizing, one can conclude that deep neural networks achieve optimal approximation properties of
all affine systems combined.

Let us finally mention that lately a very different viewpoint of expressivity was introduced in
\cite{expr_trajectory} according to so-called trajectory lengths. The standpoint taken in this work
is to measure expressivity in terms of changes of the expected length of a (non-constant) curve in the
input space as it propagates through layers of a neural network.

\subsection{Optimization}

This area aims to analyze optimization algorithms, which solve the (learning) problem in \eqref{eq:risk1},
or, more generally, \eqref{eq:risk2}. A common approach is gradient descent, since the gradient of the loss function (or optimized functional) with respect to the weight matrices and biases, i.e., the parameters of the network, can be computed exactly. This is done via backpropagation
\cite{backprop}, which is in a certain sense merely an efficient application of the chain rule. However, since the number of training samples is typically in the millions, it is computationally infeasible
to compute the gradient on each sample. Therefore, in each iteration only one or several (a batch) randomly selected gradients are computed,
leading to the algorithm of \emph{stochastic gradient descent} \cite{SGD}.

\begin{figure}[h]
\centering
\includegraphics[height=5cm]{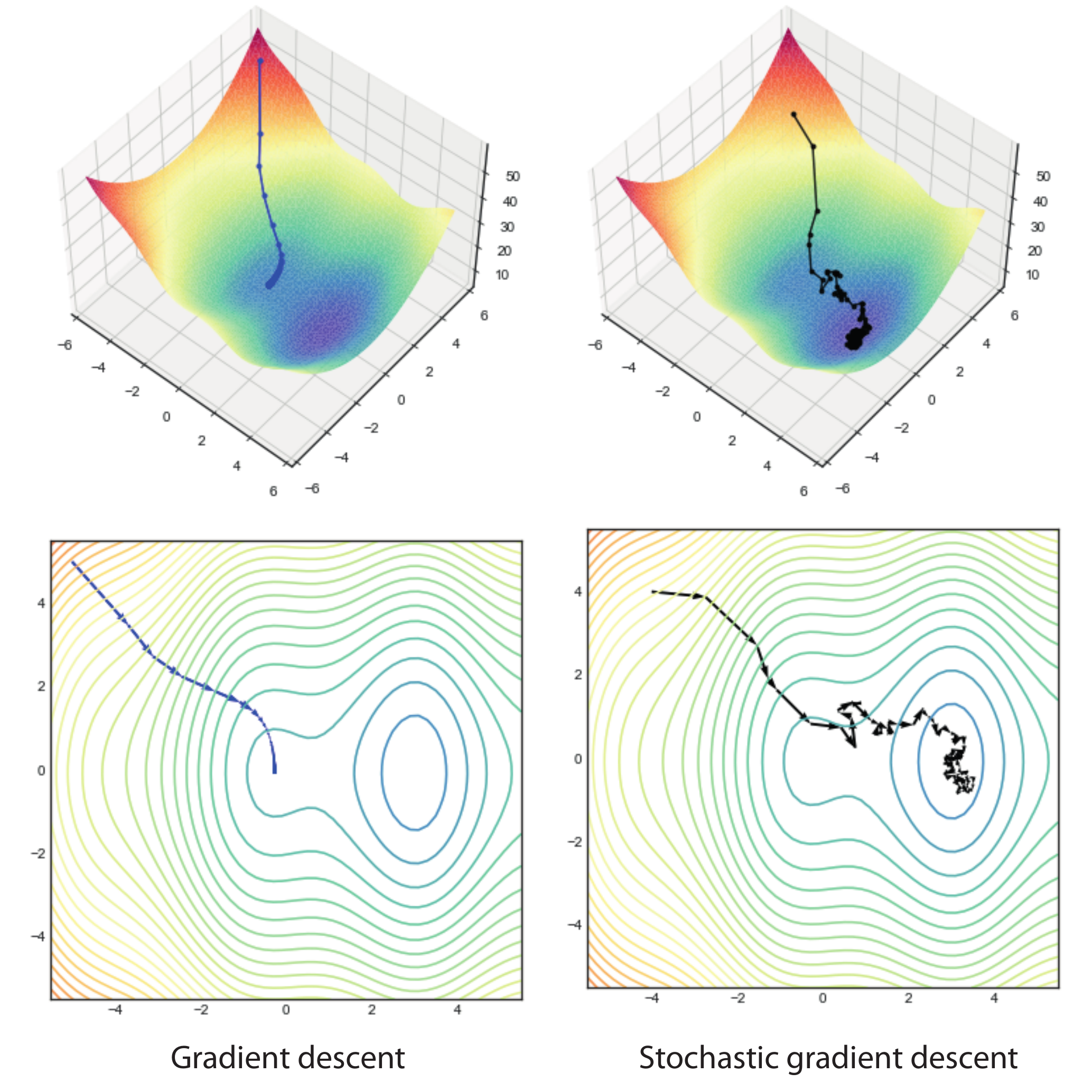}
\caption{Gradient descent versus stochastic gradient descent \cite{MMDL}}
\label{fig:33}
\end{figure}

In convex settings, guarantees for convergence of stochastic gradient descent do exist. However, in the neural network
setting, the optimization problem is non-convex, which makes it\textemdash even when using a non-random version of
gradient descent\textemdash very hard to analyze. Including randomness adds another level of difficulty as is depicted in
Figure \ref{fig:33}, where the two algorithms reach different (local) minima.

This area is by far less explored than expressivity. Most current results focus on shallow neural networks, and for
a survey, we refer to  \cite{MMDL}.

\subsection{Generalization}

This research direction is perhaps the least explored and maybe also the most difficult one, sometimes called the ``holy grail''
of understanding deep neural networks. It targets the out-of-sample error
\begin{equation}\label{eq:gener}
\sup_{\Phi \in \cNN_\theta} | \cR(\Phi)  -\widehat{\cR}(\Phi)|
\end{equation}
as described in Subsection \ref{subsec:Thread1}.

\begin{figure}[h]
\centering
\includegraphics[height=2.5cm]{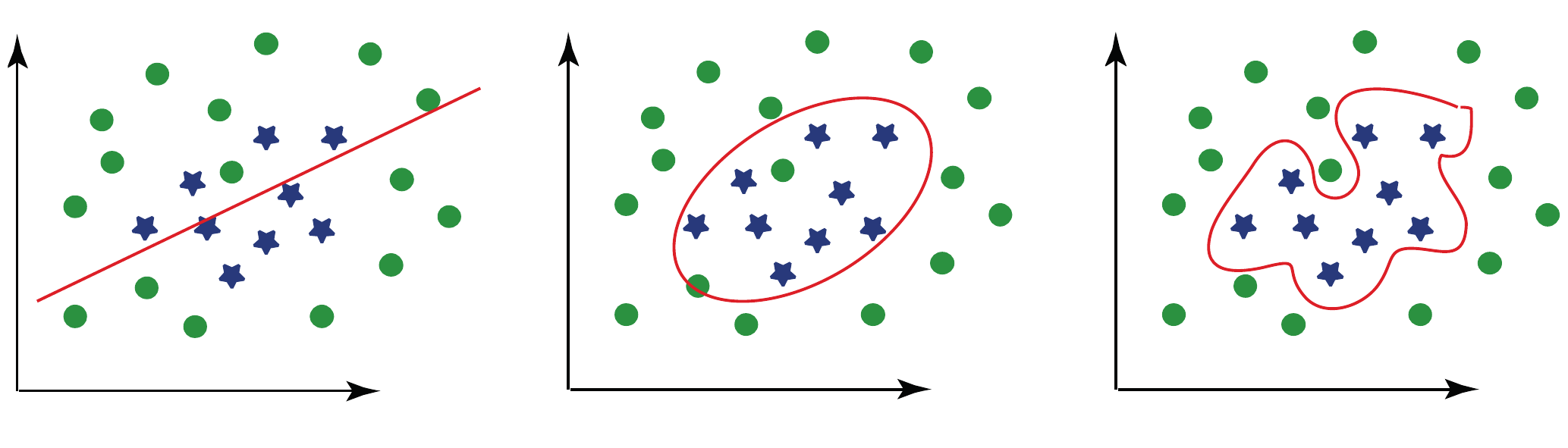}
\put(-230,0){\tiny Underfitting}
\put(-58,0){\tiny Overfitting}
\caption{Phenomenon of overfitting for the task of classification with two classes}
\label{fig:3}
\end{figure}

One of the mysteries of deep neural networks is the observation that highly overparameterized
deep neural networks in the sense of high complexity of the network do \emph{not} overfit with overfitting
referring to the problem of
fitting the training data too tightly and consequently endangering correct classification
of new data. An illustration of the phenomenon of overfitting can be found in Figure~\ref{fig:3}.

\begin{figure}[h]
\centering
\includegraphics[height=5cm]{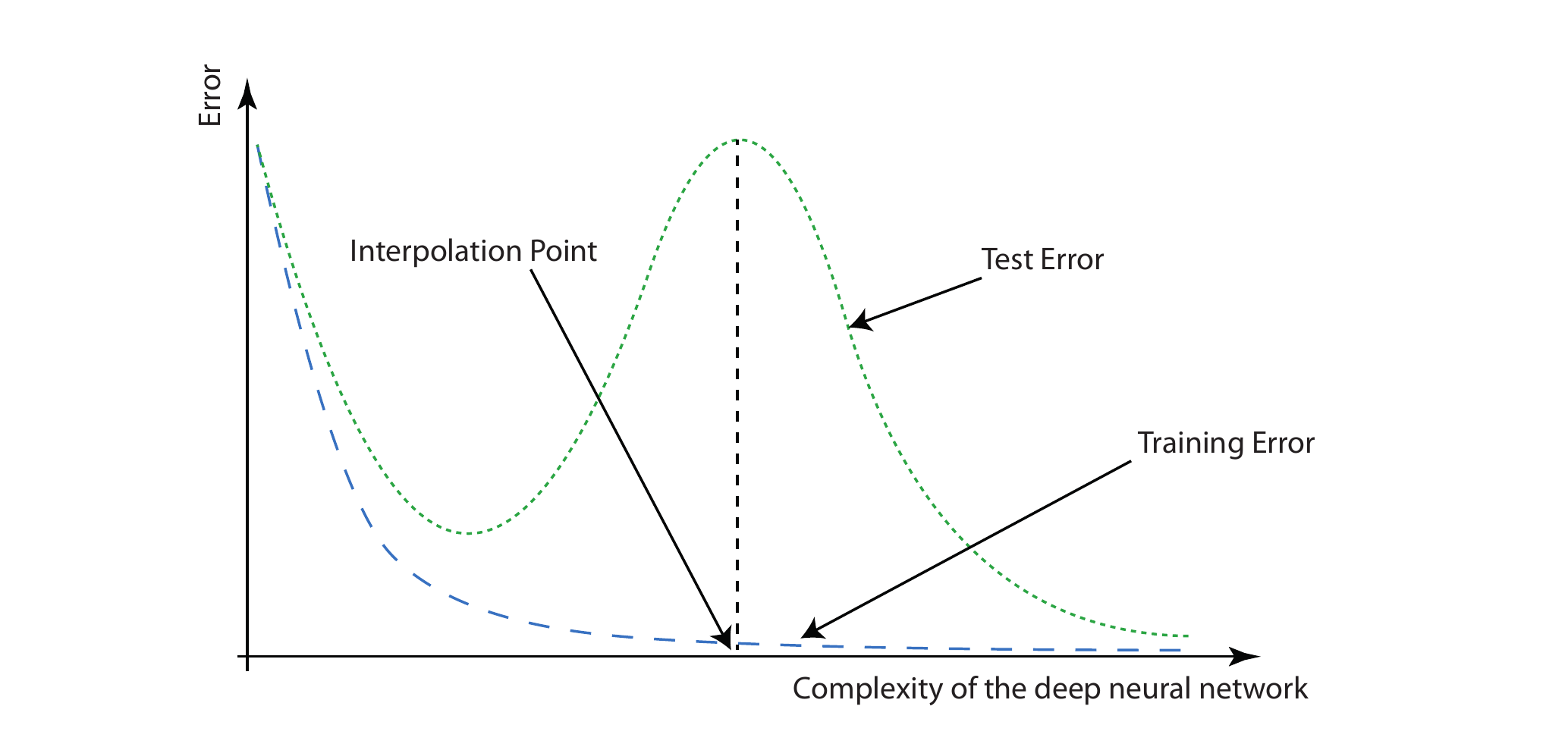}
\caption{Double descent curve}
\label{fig:4}
\end{figure}

Let us now analyze the generalization error in \eqref{eq:gener} in a bit more depth. For a large number $m$
of training samples the law of large numbers tells us that with high probability $\widehat{\cR}(\Phi) \approx \cR(\Phi)$
for each neural network $\Phi \in \cNN_\theta$. Bounding the complexity of the hypothesis class $\cNN_\theta$ by
the VC-dimension, the generalization error can be bounded with probability $1-\delta$ by
\begin{equation} \label{eq:generalization}
\sqrt{\frac{\mathrm{VCdim}(\cNN_\theta) + \log(1/\delta)}{m}}.
\end{equation}
For classes of highly over-parametrized neural networks, i.e., where $\mathrm{VCdim}(\cNN_\theta)$ is very large,
we need an enormous amount of training data to keep the generalization error under control. It is thus more than
surprising that numerical experiments show the phenomenon of a so-called \emph{double descent curve} \cite{general}.
More precisely, the test error was found to decrease after passing the interpolation point, followed by an increase consistent
with statistical learning theory (see Figure \ref{fig:4}).

\subsection{Explainability}

The area of explainability aims to ``open the black box'' of deep neural networks in the sense as to
explain decisions of trained neural networks. These explanations typically consist of providing relevance
scores for features of the input data. Most approaches focus on the task of image classification
and provide relevance scores for each pixel of the input image. One can roughly categorize the different
types of approaches into gradient-based methods \cite{Expl1}, propagation of activations in neurons \cite{Expl2},
surrogate models \cite{Expl3},  and game-theoretic approaches \cite{Expl4}.

We would now like to describe in more detail an approach which is based on information theory and
also allows an extension to different modalities such as audio data as well as analyzing the relevance
of higher-level features; for a survey paper, we refer to \cite{RDE}. This \emph{rate-distortion explanation
(RDE)} framework was introduced in 2019 and later extended by applying RDE to non-canonical input
representations.

Let now $\Phi: \R^d \to \R^{n}$ be a trained neural network, and $x \in \R^d$. The goal of RDE is to
provide an explanation for the decision $\Phi(x)$ in terms of a sparse mask $s \in \{0,1\}^d$ which highlights
the crucial input features of $x$. This mask is determined by the following optimization problem:
\[
    \min_{s\in \{0,1\}^d} \quad  \mathop{\mathbb{E}}_{v\sim\mathcal{V}} d(\Phi(x), \Phi(x\odot s + (1-s)\odot v)) \quad \mbox{subject to} \quad \norm{s}_0 \leq \ell,
\]
where $\odot$ denotes the Hadamard product, $d$ is a measure of distortion such as the $\ell_2$-distance, $\mathcal{V}$ is a distribution over input perturbations $v\in\R^d$, and $\ell\in\{1,...,d\}$ is a given sparsity level for the explanation mask $s$. The key idea is that a solution $s^*$
is a mask marking few components of the input $x$ which are sufficient to approximately retain the decision $\Phi(x)$. This viewpoint reveals the relation to rate-distortion theory, which normally focusses on lossy compression of data.

Since it is computationally infeasible to compute such a minimizer (see \cite{hard}), a relaxed optimization problem
providing continuous masks $s\in[0,1]^d$ is used in practice:
\[
\min_{s\in [0,1]^d} \quad \mathop{\mathbb{E}}_{v\sim\mathcal{V}}  d(\Phi(x), \Phi(x\odot s + (1-s)\odot v)) + \lambda \norm{s}_1,
\]
where $\lambda>0$ determines the sparsity level of the mask. The minimizer now assigns each component $x_i$ of the input\textemdash
in case of images each pixel\textemdash a relevance score $s_i\in[0,1]$.  This is typically referred to as \emph{Pixel RDE}.

\begin{figure}[h]
\centering
\includegraphics[height=7cm]{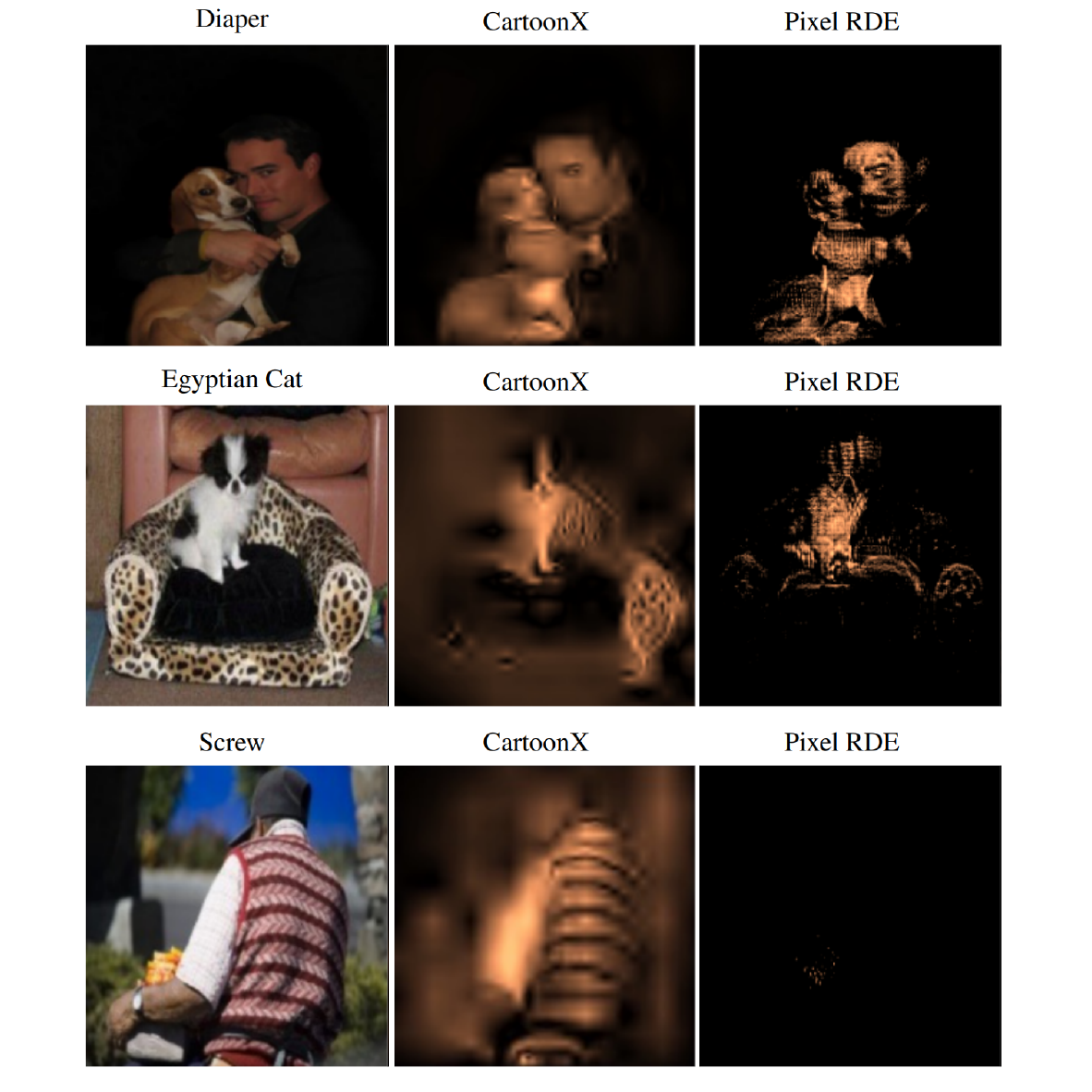}
\caption{Pixel RDE versus CartoonX for analyzing misclassifications of a deep neural network}
\label{fig:5}
\end{figure}

Extensions of the RDE-framework allow the incorporation of different distributions $\mathcal{V}$ better adapted to data distributions.
Another recent improvement was the assignment of relevance scores to higher-level features such as arising from a wavelet decomposition,
which ultimately led to the approach \emph{CartoonX}. An example of Pixel RDE versus CartoonX, which also shows the ability of the higher-level explanations
of CartoonX to give insights into what the neural network saw when misclassifying an image, is depicted in Figure \ref{fig:5}.

\section{Artificial Intelligence for Mathematical Problems}
\label{sec:AIfMP}

We now turn to the research direction of artificial intelligence for mathematical problems,
with the two most prominent problems being inverse problems and partial differential equations.
As before, we will introduce the problem settings, showcase some exemplary results, and also
discuss open problems.

\subsection{Inverse Problems}
\label{subsec:AIfMP_IP}

Methods of artificial intelligence, in particular, deep neural networks have a tremendous
impact on the area of inverse problems, as already indicated before. One current major
trend is to optimally combine classical solvers with deep learning in the sense
of taking the best out of the model- and data-world.

To introduce such results, we start by recalling some basics about solvers of inverse problems. For this, assume
that we are given an (ill-posed) inverse problem
\begin{equation} \label{eq:IP}
Kf = g,
\end{equation}
where $K : X \to Y$ is an operator and $X$ and $Y$ are, for instance, Hilbert spaces.
Drawing from the area of imaging science, examples include denoising, deblurring, or
inpainting (recovery of missing parts of an image). Most classical solvers are of the
form (which includes Tikhonov regularization)
\[
f^\alpha := \argmin_{f} \Big[ \underbrace{\|Kf-g\|^2}_{\text{Data fidelity term}} + \;  \alpha \cdot
\underbrace{\mathcal{P}(f)}_{\text{Penalty/Regularization term}} \Big],
\]
where $\mathcal{P} : X \to \R$ and $f^\alpha \in X$, $\alpha > 0$ is an approximate solution
of the inverse problem \eqref{eq:IP}. One very popular and widely applicable special case is
\emph{sparse regularization}, where $\mathcal{P}$ is chosen by
\[
\mathcal{P}(f) := \|(\langle f,\varphi_{i} \rangle)_{i \in I}\|_1
\]
and $(\varphi_{i})_{i \in I}$ is a suitably selected orthonormal basis or a frame for $X$.

We now turn to deep learning approaches to solve inverse problems, which might be categorized into
three classes:
\begin{itemize}
\item \emph{Supervised approaches.} An ad-hoc approach in this regime is given in \cite{IP2}, which first applies
a classical solver followed by a neural network to remove reconstruction artifacts. More sophisticated
approaches typically replace parts of the classical solver by a custom-build neural network  \cite{IP3} or
a network specifically trained for this task \cite{IP4}.
\item \emph{Semi-supervised approaches.} These approaches encode the regularization as a neural network
with an example being adversarial regularizers \cite{IP5}.
\item \emph{Unsupervised approaches.} A representative of this type of approaches is the technique of deep image prior \cite{IP6}.
This method interestingly shows that the structure of a generator network is sufficient to capture necessary statistics of
the data prior to any type of learning.
\end{itemize}

Aiming to illustrate the superiority of approaches from artificial intelligence for inverse problems,
we will now focus on the inverse problem of computed tomography (CT) from medical imaging. The forward
operator $K$ in this setting is the \emph{Radon transform}, defined by
\[
 \mathcal{R} f(\phi,s) = \int_{L(\phi,s)} f(x) dS(x),
\]
where $L(\phi,s) = \left\{x \in \R^2 : x_1 \cos (\phi)  + x_2 \sin (\phi) = s \right\}$, $\phi \in [-\pi/2,\pi/2)$,
and $s\in\R$. Often, only parts of the so-called sinogram $\mathcal{R} f$ can be acquired due to physical
constraints as in, for instance, electron tomography. The resulting, more difficult problem is termed
\emph{limited-angle CT}. One should notice that this problem is even harder than the problem
of low-dose CT, where not an entire block of measurements is missing, but the angular component is ``only''
undersampled.

\begin{figure}[h]
\centering
\includegraphics[width=13cm]{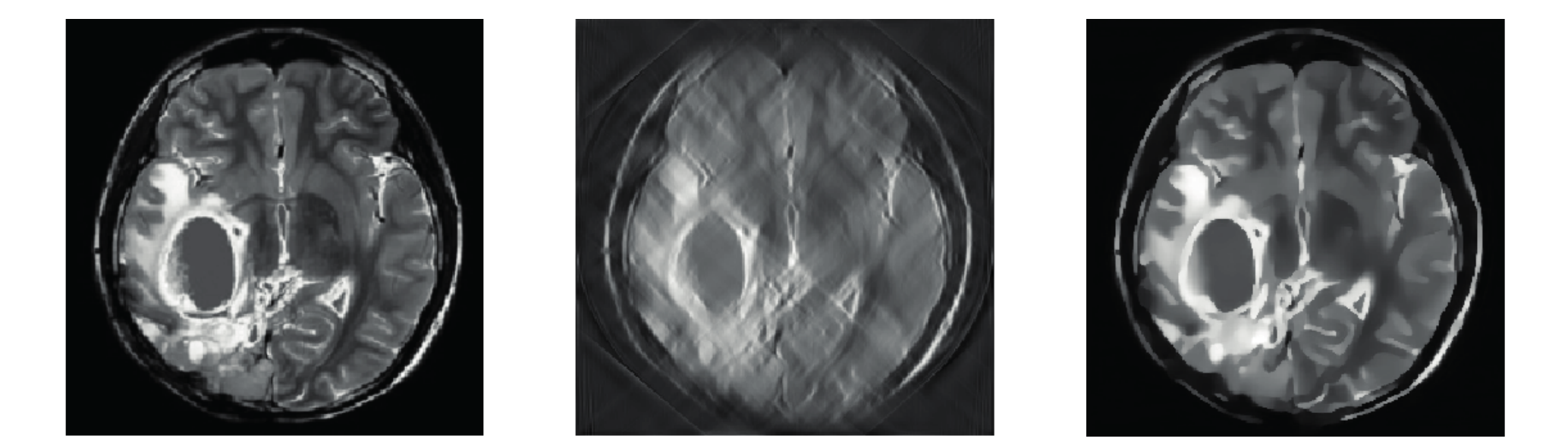}
\put(-318,-10){\tiny Original}
\put(-235,-10){\tiny Sparse Regularization with Shearlets}
\put(-114,-10){\tiny Deep Microlocal Reconstruction \cite{CT2}}
\caption{CT Reconstruction from Radon measurements with a missing angle of $40^\circ$}
\label{fig:10}
\end{figure}

The most prominent features in images $f$ are edge structures. This is also due to the fact that the
human visual system reacts most strongly to those. These structures in turn can be accurately modeled by microlocal analysis,
in particular, by the notion of wavefront sets $WF(f) \subseteq \R^2 \times \mathbb{S}$, which\textemdash coarsely
speaking\textemdash consist of singularities together with their direction. Basing in this sense the application
of a deep neural network on microlocal considerations, in particular, also using a deep learning-based wavefront
set detector \cite{AKOP19} in the regularization term, the reconstruction performance significantly outperforms
classical solvers such as  sparse regularization with shearlets (see Figure \ref{fig:10}, we also refer to \cite{CT2} for
details). Notice that this approach is of a hybrid type and takes the best out of both worlds in the sense
of combining model- and artificial intelligence-based approaches.

\begin{figure}[h]
\centering
\includegraphics[width=13cm]{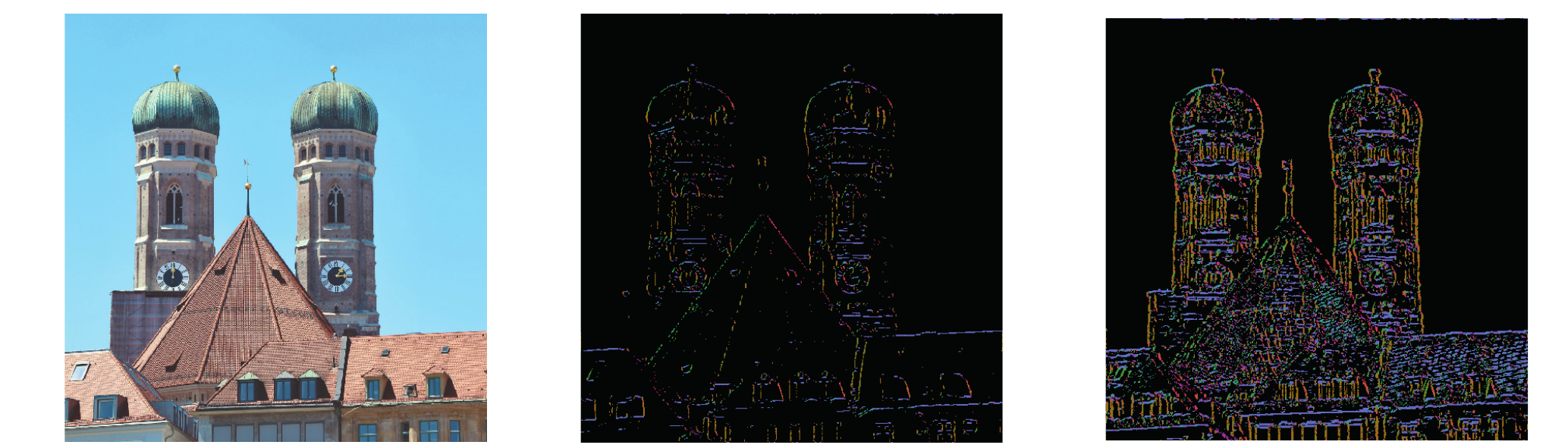}
\put(-320,-10){\tiny Original}
\put(-208,-10){\tiny CoShREM \cite{Edge2}}
\put(-75,-10){\tiny DeNSE \cite{AKOP19}}
\caption{Wavefront set detection by a model-based and a hybrid approach.}
\label{fig:11}
\end{figure}

Finally, the deep learning-based wavefront set extraction itself is yet another evidence of the improvements
on the state-of-the-art now possible by  artificial intelligence. Figure \ref{fig:11} shows a classical result from \cite{Edge2},
whereas \cite{AKOP19} uses the shearlet transform as a coarse edge detector, which is
subsequently combined with a deep neural network.

\subsection{Partial Differential Equations}
\label{subsec:AIfMP_PDE}

The second main range of mathematical problem settings, where methods from artificial intelligence
are very successfully applied to, are partial differential equations. Although the benefit of such
approaches was not initially clear, both theoretical and numerical results show their superiority
in high-dimensional regimes.

The most common approach aims to approximate the solution of a partial differential equation by a
deep neural network, which is trained according to this task by incorporating the
partial differential equation into the loss function. More precisely, given a partial differential
equation $\mathcal{L}(u) = f$, we train a neural network $\Phi$ such that
\[
\mathcal{L}(\Phi) \approx f.
\]
Since 2017, research in this general direction has significantly accelerated. Some of the highlights
are the Deep Ritz Method \cite{PDE2} and Physics Informed Neural Networks \cite{RPK19}, or a very
general approach for high-dimensional parabolic partial differential equations \cite{PDE1}.

One should note that most theoretical results in this regime are of an expressivity type and also
study the phenomenon whether and to which extent deep neural networks are able to beat the curse
of dimensionality. In the sequel, we briefly discuss one such result as an example. In addition,
notice that there already exist contributions\textemdash though very few\textemdash which analyze learning and generalization
aspects.

Let $\mathcal{L}(u_y,y) = f_y$ denote a parametric partial differential equation with $y$ being a
parameter from a high-dimensional parameter space $\mathcal{Y} \subseteq \R^p$ and $u_y$ the associated solution
in a Hilbert space $\mathcal{H}$. After a high-fidelity discretization, let
$b_y(u^{h}_y , v) = f_y(v)$ be the associated variational form with $u^h_y$
now belonging to the associated high-dimensional space $U^h$, where we set $D := \dim(U^h)$. We
moreover denote the coefficient vector of $u^h_y$ with respect to a suitable basis of $U^h$ by $\mathbf{u}_y^{\mathrm{h}}$.
Of key importance in this area is the \emph{parametric map} given by
\[
\R^p \supseteq \mathcal{Y} \ni y \; \mapsto \; \mathbf{u}_y^{\mathrm{h}} \in \R^D \quad \mbox{such that} \quad b_y(u^{h}_y , v) = f_y(v) \;\: \mbox{for all } v,
\]
which in multi-query situations such as complex design problems needs to be solved several
times. If $p$ is very large, the curse of dimensionality could lead to an exponential
computational cost.

We now aim to analyze whether the parametric map can be solved by a deep neural network,
which would provide a very efficient and flexible method, hopefully also circumventing
the curse of dimensionality in an automatic manner. From an expressivity viewpoint, one
might ask whether, for each $\epsilon > 0$, there does exist a neural network $\Phi$
such that
\begin{equation}\label{eq:pde}
\|\Phi(y) - \mathbf{u}_y^{\mathrm{h}}\| \le \epsilon \qquad \mbox{for all } y  \in \mathcal{Y}.
\end{equation}
The ability of this approach to tackle the curse of dimensionality can then be studied by
analyzing how the complexity of $\Phi$ depends on $p$ and $D$. A result of this type was
proven in \cite{PPDE1}, the essence of which we now recall.
\begin{thm}
There exists a neural network $\Phi$ which approximates the parametric map, i.e., which
satisfies \eqref{eq:pde}, and its dependence on $C(\Phi)$ on $p$ and $D$ can be (polynomially)
controlled.
\end{thm}

Analyzing the learning procedure and the generalization ability of the neural network
in this setting is currently out of reach. Aiming to still determine whether a
trained neural networks does not suffer from the curse of dimensionality as well, in
\cite{PPDE2} extensive numerical experiments were performed, which indicates that indeed
the curse of dimensionality is also beaten in practice.

\section{Conclusion: Seven Mathematical Key Problems}
\label{sec:conclusion}

Let us conclude with seven mathematical key problems of artificial intelligence as they were
stated in \cite{MMDL}. Those constitute the main obstacles in  \emph{Mathematical Foundations for Artificial Intelligence}
with its subfields expressivity, optimization, generalization, and explainability as well as in
\emph{Artificial Intelligence for Mathematical Problems} which focusses on the application
to inverse problems and partial differential equations.
\bitem
\item[(1)] What is the role of depth?
\item[(2)] Which aspects of a neural network architecture affect the performance of deep learning?
\item[(3)] Why does stochastic gradient descent converge to good local minima despite the non-convexity
of the problem?
\item[(4)] Why do large neural networks not overfit?
\item[(5)] Why do neural networks perform well in very high-dimensional environments?
\item[(6)] Which features of data are learned by deep architectures?
\item[(7)] Are neural networks capable of replacing highly specialized numerical algorithms in natural
sciences?
\eitem


\section*{Acknowledgments}
This research was partly supported by the Bavarian High-Tech Agenda, DFG-SFB/TR 109 Grant C09,
DFG-SPP 1798 Grant KU 1446/21-2, DFG-SPP 2298 Grant KU 1446/32-1, and NSF-Simons Research
Collaboration on the Mathematical and Scientific Foundations of Deep Learning (MoDL) (NSF DMS 2031985).

The author would like to thank Hector Andrade Loarca, Adalbert Fono, Stefan Kolek, Yunseok Lee,
Philipp Scholl, Mariia Seleznova, and Laura Thesing for their helpful feedback on an early version of this article.

 \end{document}